\documentclass[10pt,twocolumn,letterpaper]{article}

\usepackage[pagenumbers]{cvpr} 

\definecolor{cvprblue}{rgb}{0.21,0.49,0.74}
\usepackage[pagebackref,breaklinks,colorlinks,allcolors=cvprblue]{hyperref}

\def\paperID{6938} 
\def\confName{CVPR}
\def\confYear{2026}

\title{Proxy-Tuning: Tailoring Multimodal Autoregressive Models for Subject-Driven Image Generation}

\author{Yi Wu$^1$, Shengju Qian$^2$, Lingting Zhu$^3$, Lei Liu$^1$\thanks{Corresponding Author.} , Wandi Qiao$^1$, Ziqiang Li$^4$,\\Lequan Yu$^3$, Bin Li$^1$
\\
$^1$ University of Science and Technology of China \quad
$^2$ The Chinese University of Hong Kong \\
$^3$ The University of Hong Kong \quad
$^4$ Nanjing University of Information Science and Technology
}

\begin{document}
\maketitle
\begin{abstract}
Multimodal autoregressive (AR) models, based on next-token prediction and transformer architecture, have demonstrated remarkable capabilities in various multimodal tasks including text-to-image~(T2I) generation. Despite their strong performance in general T2I tasks, our research reveals that these models initially struggle with subject-driven image generation compared to dominant diffusion models. To address this limitation, we introduce Proxy-Tuning, leveraging diffusion models to enhance AR models' capabilities in subject-specific image generation. Our method reveals a striking weak-to-strong phenomenon: fine-tuned AR models consistently outperform their diffusion model supervisors in both subject fidelity and prompt adherence. We analyze this performance shift and identify scenarios where AR models excel, particularly in multi-subject compositions and contextual understanding. This work not only demonstrates impressive results in subject-driven AR image generation, but also unveils the potential of weak-to-strong generalization in the image generation domain, contributing to a deeper understanding of different architectures' strengths and limitations. 
\end{abstract}    
\section{Introduction}
\label{sec:intro}

Recently, multimodal autoregressive (AR) models~\cite{chern2024anole,team2024chameleon,liu2024lumina,ge2024seed,wang2024emu3}, based on next-token prediction and transformer architecture~\cite{vaswani2017attention}, have demonstrated remarkable capabilities across various multimodal tasks, including text-to-image~(T2I) generation. These models have shown comparable performance to large diffusion models~\cite{rombach2022high,podell2023SDXL,esser2024scaling} in general image generation tasks. However, the application of AR models to subject-driven image generation, a task of significant user interest~\cite{ruiz2023dreambooth}, remains relatively unexplored compared to the extensive research on diffusion models in this domain~\cite{ruiz2023dreambooth,chen2023disenbooth,alaluf2023neural,gal2022image,ye2023ip,li2024photomaker,wu2024infinite}.


Subject-driven image generation~\cite{ruiz2023dreambooth,gal2022image,kumari2023multi} requires a model to learn the visual appearance of a specific subject from a small set of reference images and then generate novel images of that subject based on user prompts. While diffusion models~\cite{ho2020denoising, rombach2022high} have been successfully adapted for this task using methods like DreamBooth~\cite{ruiz2023dreambooth}, our initial attempts to tailor AR models for subject-driven generation revealed significant challenges.


We explore two primary approaches to adapt AR models for this task: parameter-efficient fine-tuning using LoRA~\cite{hu2021lora} and end-to-end fine-tuning. In both cases, we bound the given subject to a pre-defined token (\eg, S*) for use during inference. However, our experiments uncover limitations in both approaches. LoRA-based fine-tuning struggles to capture the specific subject's appearance accurately, while end-to-end fine-tuning leads to diminished text fidelity in the generated images and compromises the model's original capabilities.


To address these limitations, we introduce Proxy-Tuning, a novel method that leverages diffusion models to enhance AR models' capabilities in subject-specific image generation. Our approach involves first fine-tuning a relatively weaker diffusion model on the reference images of the specific subject, and then uses the text-image pairs generated by this fine-tuned diffusion model to supervise the fine-tuning of more powerful AR models.


Surprisingly, we discover a striking weak-to-strong generalization~\cite{burns2023weak} phenomenon: fine-tuned AR models consistently outperform their diffusion model supervisors in both subject fidelity and prompt adherence. This trend holds true across different combinations and architectures of teacher (diffusion) and student (AR) models. Further analysis reveals scenarios where AR models particularly excel, \eg, multi-subject compositions and contextual understanding.

Our work demonstrates impressive results in subject-driven image generation and unveils the potential of weak-to-strong generalization in this domain. This phenomenon highlights the unique capabilities of AR models, particularly their ability to generalize and improve upon the knowledge distilled from diffusion models. The success of Proxy-Tuning suggests that AR models possess a remarkable capacity to synthesize and extend learned information, potentially due to their sequential processing nature and large-scale pretraining on diverse data. By bridging the gap between diffusion and AR models, our work not only advances the field of subject-driven image generation but also opens new avenues for leveraging AR models in diverse multimodal applications.

\section{Related Work}
\label{sec:related}
\subsection{Subject-Driven Image Generation}
Subject-driven image generation is a task that focuses on generating images related to a specific subject, where the model learns from a few reference images of the subject and combines with given prompts to create novel images of that subject. Currently, the existing methods regarding subject-driven image generation are mainly divided into two categories. The first category is tuning-based methods~\cite{ruiz2023dreambooth,chen2023disenbooth,alaluf2023neural,gal2022image,kumari2023multi}, which require fine-tuning the model for each specific target. Specifically, DreamBooth~\cite{ruiz2023dreambooth} binds the given subject to a specially pre-defined token (\eg, S*) and introduces a class-specific prior preservation loss to preserve the original knowledge of the pretrained diffusion model. Disenbooth~\cite{chen2023disenbooth} proposes to enhance the disentanglement of the specific subject and background features during the generation process, aiming to improve the quality and flexibility of the generated images. NeTI~\cite{alaluf2023neural} focuses on improving the neural network architectures used in subject-driven image generation to achieve better visual fidelity and semantic consistency. The another category is tuning-free methods~\cite{ye2023ip,li2024photomaker,wu2024infinite,xiao2024fastcomposer,wang2024instantid,ma2024subject,shi2024instantbooth,wei2023elite}. Among these methods, IP-Adapter~\cite{ye2023ip} introduces a decoupled cross-attention mechanism to capture and fuse the image embedding input and text embedding input to guide the process of image generation. PhotoMaker~\cite{li2024photomaker} extracts image features by using a pretrained CLIP image encoder and projects the image features to the text latent space through a lightweight mapper. Eventually, these image features are concatenated with the text features and then input into the model. Similarly, ELITE~\cite{wei2023elite} also extracts image features by using a pretrained encoder, and incorporates the image features into the text latent space and the cross-attention module respectively by means of global mapping and local mapping.

\subsection{Multimodal Autoregressive Models}
Multimodal autoregressive (AR) models~\cite{sun2024autoregressive,team2024chameleon,liu2024lumina,wang2024emu3,ge2024seed,chern2024anole}, based on next-token prediction and transformer architecture, have shown impressive abilities in various multimodal tasks, especially in the realm of text-to-image~(T2I) generation. LlamaGen~\cite{sun2024autoregressive} explores the application of the "next-token prediction" paradigm from large language models to visual generation. By scaling the widely used autoregressive Llama backbones, it attains excellent image generation performance. Similarly, Chameleon~\cite{team2024chameleon} also adopts the autoregressive generation approach, capable of transforming images and text into discrete tokens for processing, showing excellent performances in various tasks such as visual question answering and image generation. Based on Chameleon, Lumina-mGPT~\cite{liu2024lumina} enhances the training datasets and achieves flexible text-to-image generation and controllable generation, also capable of handling vision recognition tasks. Moreover, Emu3~\cite{wang2024emu3} further includes the video datasets in the training datasets and trains the transformer model only with next-token prediction by encoding images, texts and videos into a discrete space and jointly training on multimodal mixed sequences.
\section{Analysis: Subject-Driven AR Tuning}
\label{sec:analysis}

\begin{figure}[t]
  \includegraphics[width=\columnwidth]{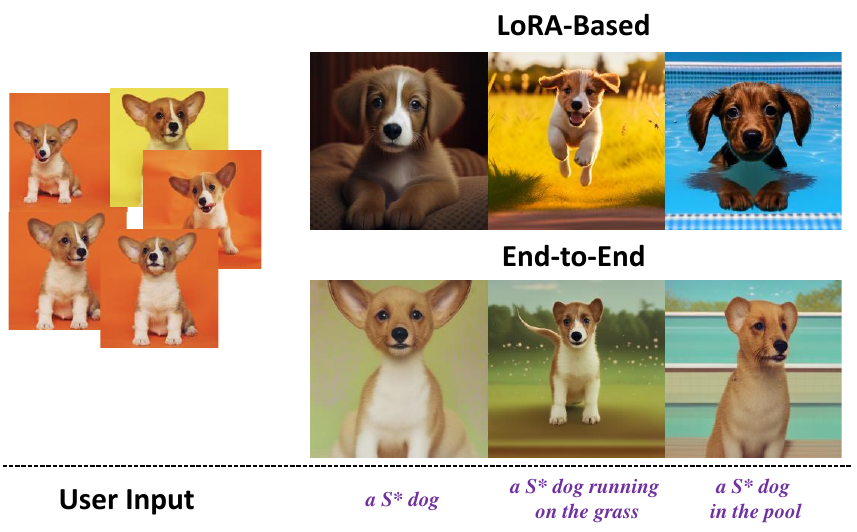}
  \caption{Visualizations of parameter-efficient~(LoRA) and end-to-end direct subject fine-tuning on AR model.}
  \label{fig:analysis}
\end{figure}

\begin{table}
  \centering
  \resizebox{.47\textwidth}{!}{
  \begin{tabular}{lllll}
    \hline
     & \textbf{Model} & \textbf{CLIP-I} & \textbf{CLIP-T} & \textbf{DINO} \\
    \hline
    LoRA & SDXL~\cite{podell2023SDXL}  &0.8002&0.3225&0.7272 \\ \hline
    End-to-End & Lumina-mGPT~\cite{liu2024lumina}    & 0.6974 &0.2652      &0.5338       \\
    LoRA & Lumina-mGPT~\cite{liu2024lumina}    & 0.6752  & 0.2956 & 0.5088            \\
    \hline
  \end{tabular}
  }
  \caption{Quantitative evaluation of DreamBooth~\cite{ruiz2023dreambooth} subject-driven fine-tuning on AR model. Details about the metrics are provided in Section~\ref{sec:dataset}.
  }
  \label{tab:analysis}
\end{table}

\begin{figure*}[t]
  \includegraphics[width=1\linewidth]{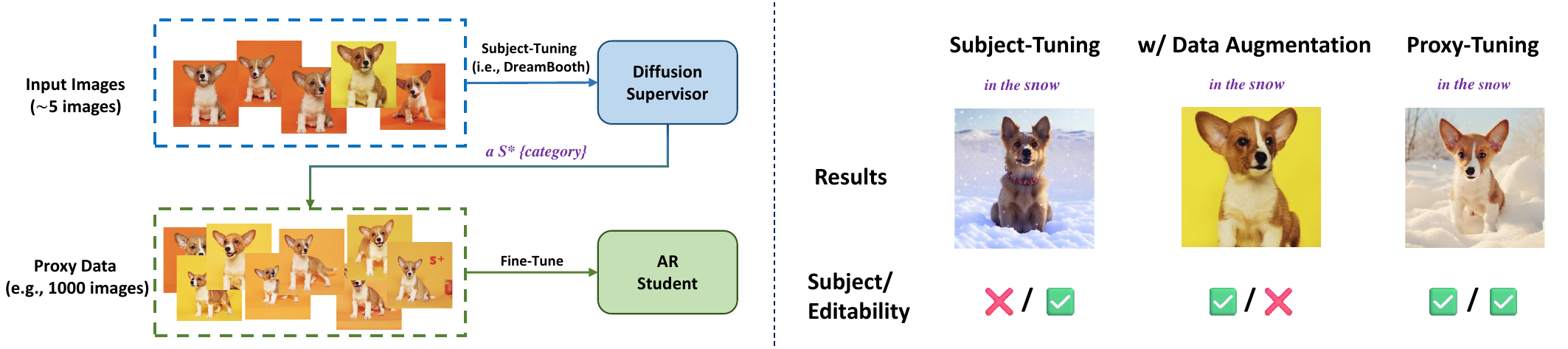}
  \caption {Framework of Proxy-Tuning. Provided with a limited number of images of a particular target, we initially conduct subject-tuning (\ie, DreamBooth) on the diffusion model. Then the diffusion model is employed as a supervisor to supervise the fine-tuning of the AR model. Direct subject-tuning of the AR model results in an unsatisfactory acquisition of the target appearance, and the utilization of data augmentations in subject-tuning manifests a deficiency in semantic editability (as discussed in Section~\ref{sec:ablation}). Conversely, Proxy-Tuning effectively captures the target appearance and simultaneously showcases excellent semantic editability.}
  \label{fig:method}
\end{figure*}

To explore subject-driven image generation capabilities in AR models, we initially explored the settings in DreamBooth~\cite{ruiz2023dreambooth} of fine-tuning models with few subject images.  Our investigation focused on two primary scenarios: end-to-end fine-tuning of the entire model and parameter-efficient tuning using LoRA~\cite{hu2021lora}. Following the DreamBooth paradigm, we associate the target subject with a designated token (e.g., "S*") for inference purposes. 

Our experimental results, as illustrated in Figure~\ref{fig:analysis} and quantified in Table~\ref{tab:analysis}, reveal significant limitations in both approaches:

\begin{itemize}
\setlength{\itemsep}{0pt}
\setlength{\parsep}{0pt}
\setlength{\parskip}{0pt}
\item \textbf{LoRA fine-tuning} maintains reasonable semantic consistency in generated images but struggles to effectively capture and reproduce specific subject features.
\item \textbf{End-to-end fine-tuning} performs poorly on both aspects: it neither adequately captures the target subject's features nor maintains semantic consistency in the generated images, with particularly severe degradation in AR models' semantic following capabilities.
\end{itemize}

Notably, these challenges are less prominent in diffusion models under similar few-shot subject-driven settings~\cite{ruiz2023dreambooth}. This disparity can be attributed to the inherent characteristics of AR models: they generate images through sequential token predictions where each token heavily depends on previous ones. This autoregressive nature makes them particularly sensitive to parameter updates when trained on very few subject images, as even slight perturbation in token prediction can propagate and amplify throughout the generation sequence. In contrast, diffusion models' parallel denoising process demonstrates greater robustness to few-shot fine-tuning, better preserving both subject fidelity and semantic consistency.

These findings underscore the fundamental challenges in adapting AR models for subject-driven image generation through direct fine-tuning approaches, necessitating the development of more sophisticated methods that can better preserve the model's original capabilities while effectively learning subject-specific features.


\section{Method: Proxy-Tuning}
\label{sec:method}

Our objective is to enhance pretrained AR models for subject-driven image generation while preserving their original semantic understanding capabilities. Specifically, we aim to enable these models to accurately learn features of specific subjects and seamlessly integrate these learned features with text inputs to generate novel, subject-related images. To address the challenges identified in direct subject fine-tuning approaches, we propose Proxy-Tuning, a novel method that effectively extends the model's subject-driven generation capabilities.


As illustrated in Figure~\ref{fig:method}, Proxy-Tuning operates through a three-stage process:
\begin{itemize}
\setlength{\itemsep}{0pt}
\setlength{\parsep}{0pt}
\setlength{\parskip}{0pt}
\item \textbf{Diffusion Supervisor Learning:} We fine-tune a diffusion model with a parameter-efficient method, \ie, LoRA, to capture the specific subject's appearance. The subject is bound to a predefined token (e.g., "S*") for controlled generation.
\item \textbf{Proxy Data Synthesis:} Using the fine-tuned diffusion model, we generate a diverse dataset using prompts in the format "a S* \{category\}". These synthesized images serve as proxy training data for the AR model.
\item \textbf{AR Student Learning:} We fine-tune the AR model using LoRA on the synthesized dataset, enabling it to learn the subject's features while maintaining its broader semantic capabilities.
\end{itemize}

Notably, we discover a compelling weak-to-strong phenomenon in image generation, similar to observations in natural language processing~\cite{burns2023weak}. Our findings show that AR models, when trained under the supervision of a weak diffusion model, consistently outperform their supervisors in both subject fidelity and prompt adherence. This is the first demonstration of such a phenomenon in the context of multimodal AR models for image generation.


\begin{table*}[ht]
  \resizebox{\textwidth}{!}{
  \begin{tabular}{lccc}
    \hline
    Supervisor: SDXL & CLIP-I & CLIP-T & DINO \\
    \hline
     \textbf{Supervisor} & & & \\
    \hline
    SDXL     &0.8002  &\textbf{0.3225}      &0.7272       \\
        \hline
     \textbf{Student} & & & \\
    \hline
    Lumina-mGPT     &0.6721  & 0.3101 & 0.4504              \\
    \textbf{Lumina-mGPT w/ Proxy-Tuning}     &\underline{0.8074}  & \underline{0.3118} & \textbf{0.7834}            \\
    LlamaGen  &0.6752&0.2956&0.5088 \\
    \textbf{LlamaGen w/ Proxy-Tuning}  &\textbf{0.8152}&0.2772&\underline{0.7436} \\
    \hline
  \end{tabular}
  \begin{tabular}{lccc}
    \hline
    Supervisor: SD3 & CLIP-I & CLIP-T & DINO \\
    \hline
     \textbf{Supervisor} & & & \\
    \hline
    SD3     &0.7822  &\textbf{0.3247}      &0.7069       \\
    \hline
     \textbf{Student} & & & \\
    \hline
    Lumina-mGPT     &0.6721  & 0.3101 & 0.4504              \\
    \textbf{Lumina-mGPT w/ Proxy-Tuning}    &\underline{0.7977}  & \underline{0.3167} & \textbf{0.7551}            \\
    LlamaGen  &0.6752&0.2956&0.5088 \\
    \textbf{LlamaGen w/ Proxy-Tuning} &\textbf{0.8105}&0.2839&\underline{0.7381} \\
    \hline
  \end{tabular}
  }
  \caption{Quantitative evaluation with SDXL and SD3 as the supervisor models. 
  The best results are in \textbf{bold}, and the second-best results are \underline{underline}. Lumina-mGPT and LlamaGen are the models obtained by directly fine-tuning Lumina-mGPT and and LlamaGen with LoRA using the original DreamBooth dataset that contains only a few target images. Lumina-mGPT w/ Proxy-Tuning and LlamaGen w/ Proxy-Tuning are the models obtained through our Proxy-Tuning method.}
  \label{tab:comparison}
\end{table*}

\begin{table*}[ht]
\resizebox{\textwidth}{!}{
  \begin{tabular}{lccc}
    \hline
   Supervisor: SD3.5  & CLIP-I & CLIP-T & DINO \\
    \hline
     \textbf{Supervisor} & & & \\
    \hline
    SD3.5     &0.7891  &\textbf{0.3271}      &0.7128       \\ \hline
     \textbf{Student} & & & \\
    \hline
    Lumina-mGPT     &0.6721  & 0.3101 & 0.4504              \\
    \textbf{Lumina-mGPT w/ Proxy-Tuning}     &\underline{0.8091}  & \underline{0.3123} & \textbf{0.7523}            \\
    LlamaGen  &0.6752&0.2956&0.5088 \\
    \textbf{LlamaGen w/ Proxy-Tuning} &\textbf{0.8179}&0.2770&\underline{0.7407} \\
    \hline
  \end{tabular}
  \begin{tabular}{lccc}
    \hline
    Supervisor: FLUX  & CLIP-I & CLIP-T & DINO \\
    \hline
     \textbf{Supervisor} & & & \\
    \hline
    FLUX     &\underline{0.8023}  &\underline{0.3144}      &\underline{0.7324}       \\
    \hline
     \textbf{Student} & & & \\
    \hline
    Lumina-mGPT     &0.6721  & 0.3101 & 0.4504              \\
    \textbf{Lumina-mGPT w/ Proxy-Tuning}     &\textbf{0.8098}  & \textbf{0.3155} & \textbf{0.7564}            \\
    LlamaGen  &0.6752&0.2956&0.5088 \\
    \textbf{LlamaGen w/ Proxy-Tuning} &0.7974&0.2800&0.7286 \\
    \hline
  \end{tabular}
  }
  \caption{Quantitative evaluation with SD3.5 and FLUX as the supervisor models. 
   We present the extended results of Table~\ref{tab:comparison}, where the supervisor models are different.
  }
  \label{tab:comparison_1}
\end{table*}

\section{Experiments}
\label{sec:experiments}

This section presents our experimental setup and results. We begin by describing datasets and evaluation metrics in Section~\ref{sec:dataset}. Section~\ref{sec:model} details the architecture and training specifications of both the diffusion supervisor and the autoregressive student model. In Section~\ref{sec:comparison}, we demonstrate the effectiveness of our Proxy-Tuning through comprehensive quantitative and qualitative analysis.

\subsection{Dataset and Evaluation}
\label{sec:dataset}
\noindent\textbf{Dataset.} Our experiments utilize the subject-driven image generation dataset introduced in DreamBooth~\cite{ruiz2023dreambooth}. We curate a diverse training dataset comprising 9 subjects: 4 live subjects (\eg, dog, cat) and 5 inanimate objects. Following DreamBooth~\cite{ruiz2023dreambooth}, to ensure robust quantitative evaluation, we generate test images using 25 distinct prompts for each subject category. During testing, we produce 4 images per prompt for each subject, resulting in a comprehensive evaluation set of 225 images.

\noindent\textbf{Evaluation Metrics.} Subject-driven image generation requires evaluation along two critical dimensions: subject fidelity and prompt adherence. For subject fidelity, we employ two metrics: CLIP-I and DINO~\cite{caron2021emerging}. CLIP-I computes the cosine similarity between CLIP~\cite{radford2021learning} image embeddings of real and generated images, while DINO calculates the same similarity using embeddings from ViT-S/16~\cite{dosovitskiy2020image}. For prompt fidelity, we use CLIP-T to measure the cosine similarity between CLIP image embeddings of generated images and text embeddings of their corresponding prompts.

\subsection{Model Selection}
\label{sec:model}
\noindent\textbf{Models.} To validate the generalization capability of Proxy-Tuning across different model scales and architectures, we experiment with various diffusion supervisors and AR students for subject-driven image generation. For diffusion supervisors, we select models with different architectures and parameter scales but all trained using diffusion-based methods: SDXL~\cite{podell2023SDXL} (U-Net, 2.6B parameters), SD3 Medium~\cite{esser2024scaling} (DiT, 2B parameters), SD3.5 Large~\cite{esser2024scaling} (DiT, 8B parameters), and FLUX.1 [dev] (DiT, 12B parameters). For AR students, we employ two models trained with autoregressive objectives: LlamaGen-XL from LlamaGen~\cite{sun2024autoregressive} (0.775B parameters) and FP-SFT@768 from Lumina-mGPT~\cite{liu2024lumina} (7B parameters). Unless otherwise specified, all experiments use parameter-efficient training with LoRA rather than full model fine-tuning.

\begin{figure*}[t]
  \includegraphics[width=0.95\linewidth]{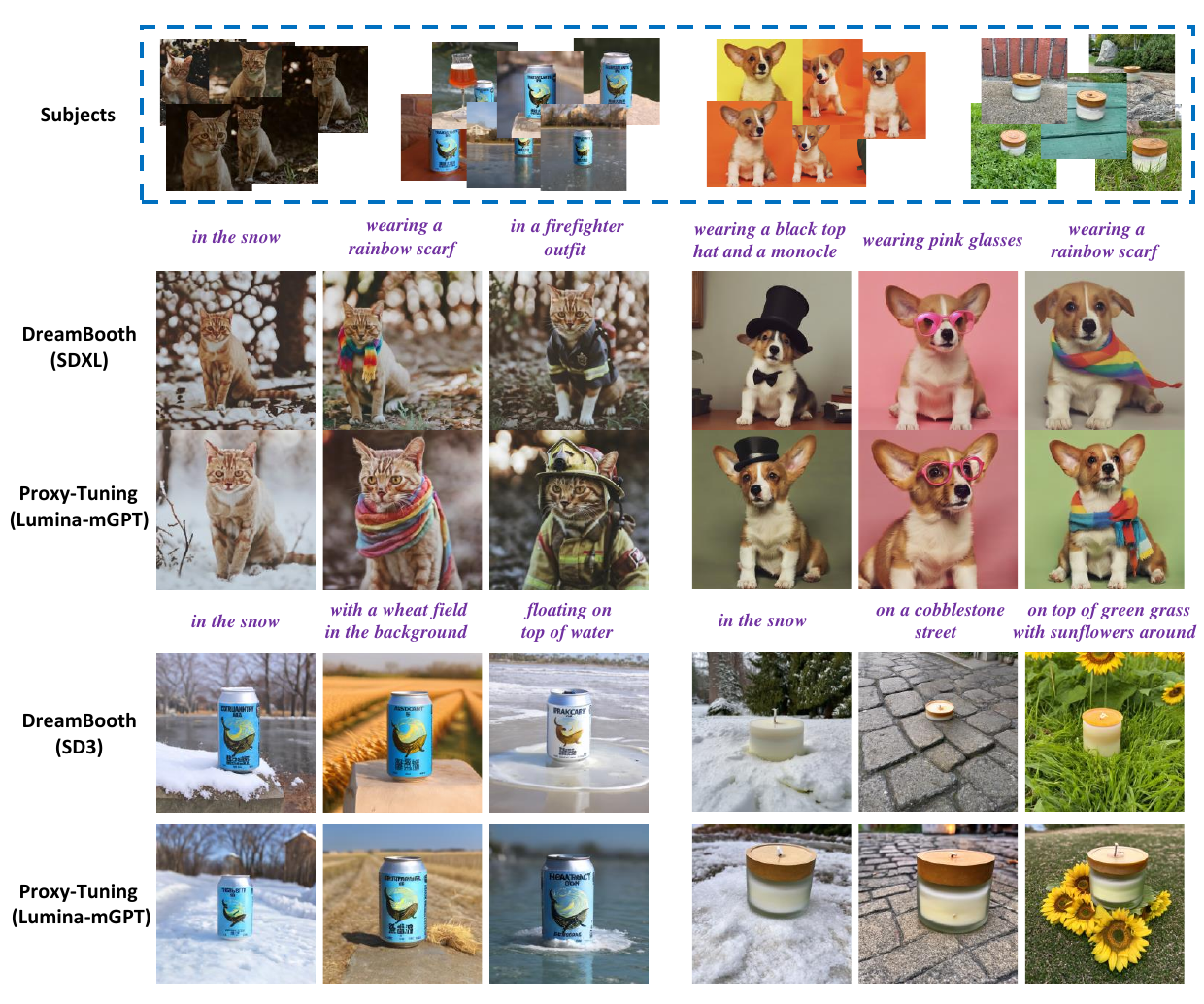}
  \caption {Visualization of our Proxy-Tuning method. We use SDXL and SD3 respectively as the weak supervisors to supervise and fine-tune Lumina-mGPT. Eventually, the subject appearance (the images in the first row) learned by the fine-tuned Lumina-mGPT (the images in the third row and the images in the fifth row) is even better than that of their weak supervisors (the images in the second row and the images in the fourth row).}
  \label{fig:comparison}
\end{figure*}

\subsection{Comparisons}
\label{sec:comparison}
\noindent\textbf{Direct Subject Tuning.} We first investigate the effectiveness of direct subject fine-tuning on state-of-the-art AR models using the DreamBooth dataset. Both Lumina-mGPT (7B parameters) and LlamaGen (775M parameters), despite their different scales and training datasets, fail to capture subject-specific characteristics through either parameter-efficient or end-to-end fine-tuning. The quantitative and qualitative results are shown in Table~\ref{tab:analysis} and Figure~\ref{fig:analysis}, respectively.

\noindent\textbf{Proxy-Tuning of AR Models.} We evaluate our Proxy-Tuning method using various diffusion supervisors: SDXL (U-Net, 2.6B parameters), SD3 Medium (DiT, 2B parameters), SD3.5 Large (DiT, 8B parameters), and FLUX.1 [dev] (DiT, 12B parameters). Our approach first fine-tunes these diffusion models on specific targets, then uses their generated images to supervise the AR models. The results in Table~\ref{tab:comparison}, Table~\ref{tab:comparison_1}, and Figure~\ref{fig:comparison} demonstrate that AR models not only effectively learn target characteristics but also surpass their diffusion supervisors in terms of CLIP-I and DINO metrics, regardless of the supervisor's architecture. We term this surprising improvement the weak-to-strong generalization phenomenon.

\noindent\textbf{Proxy-Tuning of Diffusion Models.} To investigate whether the weak-to-strong generalization is unique to AR models, we apply Proxy-Tuning to diffusion models. Specifically, this involves having diffusion supervisors supervise the fine-tuning process of diffusion models. By including more proxy subject samples and training iterations, experiments in Table~\ref{tab:proxy_diffusion} consistently show that diffusion models perform worse than their supervisors in subject-driven image generation, which confirms that the weak-to-strong generalization phenomenon is specific to AR models.

\noindent\textbf{Findings.} Our comprehensive experiments with various AR students and architecturally diverse diffusion supervisors demonstrate that: (1) Proxy-Tuning enables effective subject-driven image generation in AR models regardless of supervisor architecture, and (2) AR models exhibit weak-to-strong generalization, consistently outperforming their diffusion supervisors. This phenomenon is unique to AR models and does not occur when applying Proxy-Tuning to diffusion models. 

\noindent\textbf{Why does the weak-to-strong generalization only occur in AR models but not in diffusion models?} According to previous research~\cite{parihar2024balancing,yue2024few,ning2023elucidating}, diffusion models often exhibit biased learning towards certain attributes in training datasets, and during inference, their progressive denoising process is susceptible to exposure bias\cite{ning2023elucidating}, causing generated images to drift away from the target data distribution. These issues manifest themselves in subject-driven image generation tasks as diffusion models struggle to consistently reproduce subjects' appearance attributes, leading to variance in generated images. In this case, when using Proxy-Tuning in diffusion models, these issues are exacerbated, leading to significant bias in the appearance of the subjects (as shown in Table~\ref{tab:proxy_diffusion}). Compared to diffusion models, AR models encode the target subject's image into a discrete token distribution, which is divided into two parts: a primary distribution composed of tokens representing the local appearance of the target subject, and a minor distribution formed by biased tokens introduced through diffusion models. During next-token prediction training, the AR model tends to fit the primary distribution (which captures the local appearance of the target subject) while filtering out the minor distribution formed by biased tokens. This allows the AR model to learn a more accurate representation of the target subject's appearance.

\begin{table}
  \resizebox{.48\textwidth}{!}{
  \begin{tabular}{lccc}
    \hline
    & CLIP-I & CLIP-T & DINO \\
    \hline
     \textbf{Student} & & & \\
    \hline
    SDXL & 0.7941 ($\downarrow$0.61\%) & 0.3168 ($\downarrow$5.70\%) & 0.6890 ($\downarrow$3.82\%) \\
    SD3 & 0.7596 ($\downarrow$2.26\%) &\bf 0.3248 ($\uparrow$0.01\%) & 0.6569 ($\downarrow$5.00\%)  \\
    SD3.5 & 0.7619 ($\downarrow$2.72\%) &\bf 0.3313 ($\uparrow$0.42\%) & 0.6607 ($\downarrow$5.21\%) \\
    FLUX  & 0.7962 ($\downarrow$0.61\%)  & 0.3088 ($\downarrow$0.56\%)  & 0.6939 ($\downarrow$3.85\%)  \\
    \hline
  \end{tabular}
  }
  \caption{Quantitative evaluation with diffusion models being the student. $\downarrow$ and $\uparrow$ denote the deterioration and improvement over their DreamBooth baselines.
  }
  \label{tab:proxy_diffusion}
\end{table}

\subsection{Multiple Subjects Personalization}
Given that AR models have demonstrated superior scalability to diffusion models~\cite{tian2024visual}, we investigate their capability in multi-subject personalization. While diffusion models typically require separate model instances or parameters for different subjects, we hypothesize that AR models can effectively learn multiple subjects in a single pass through our Proxy-Tuning framework.

To test this hypothesis, we train a single AR model (LlamaGen) to simultaneously learn multiple subjects from the proxy data of separate diffusion supervisors. Each diffusion supervisor is first individually fine-tuned via LoRA to learn a specific subject. We then train the AR student to generate four randomly selected subjects using distinct prompts (denoted as $S^*_1$, $S^*_2$, etc.) designed to elicit subject-specific characteristics.

\begin{figure*}[t]
  \includegraphics[width=1\linewidth]{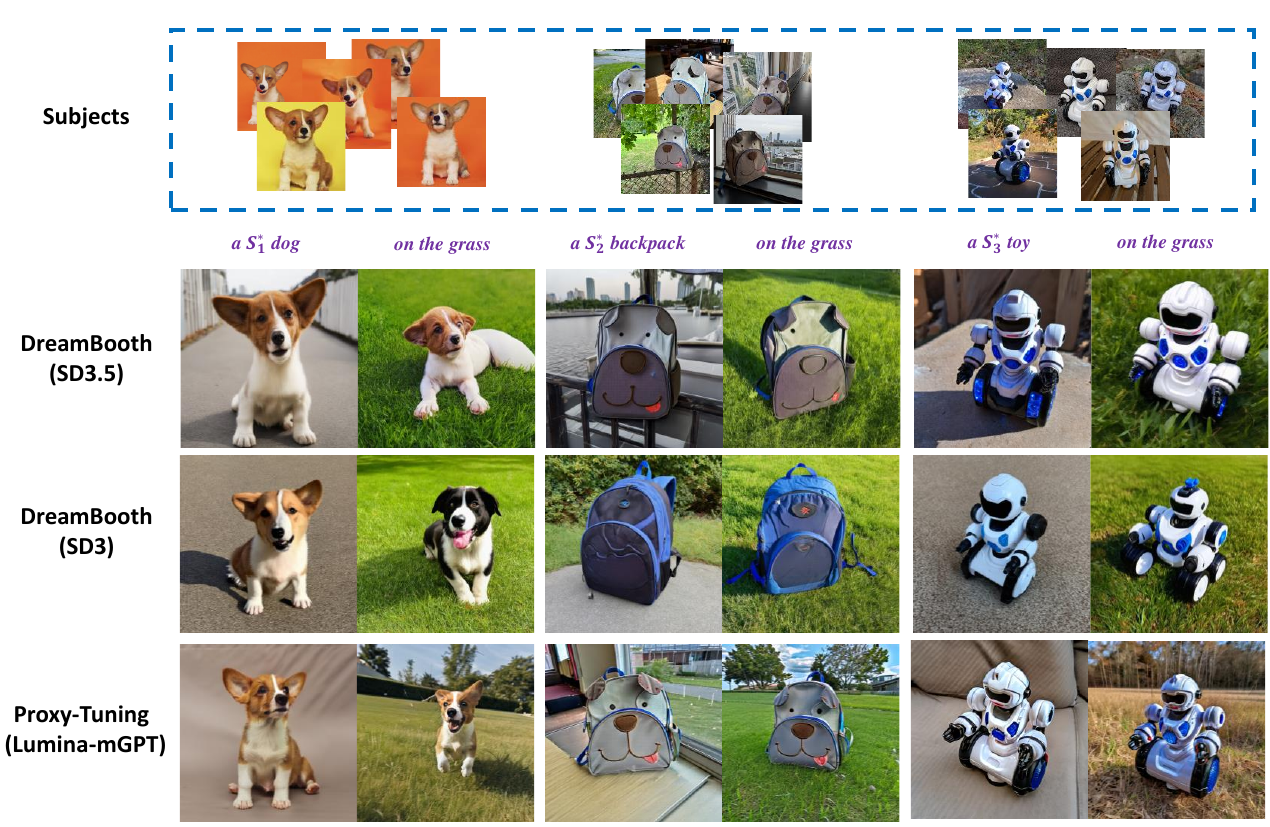}
  \caption{Visualization of the \textbf{multiple subjects} personalization. We fine-tune SD3 and SD3.5 to learn multiple subjects simultaneously and employ Proxy-Tuning on Lumina-mGPT with the supervision of SDXL to learn multiple subjects simultaneously.}
  \label{fig:multi}
\end{figure*}

The quantitative results in Table~\ref{tab:multi} reveal a striking finding: our AR model, even when learning multiple subjects simultaneously, matches the subject fidelity of individual diffusion supervisors that are specialized for single subjects. For comparison, we also train diffusion models to learn multiple subjects simultaneously. As shown in Figure~\ref{fig:multi}, while our AR model maintains distinct characteristics for each subject, the diffusion model exhibits severe subject mixing and degraded image quality, often confusing or blending features from different subjects. This visual evidence, along with the quantitative results, demonstrates the superior capability of AR models in handling multi-subject personalization tasks.

\begin{table}
  \resizebox{.47\textwidth}{!}{
  \begin{tabular}{lcccc}
    \hline
    & Separate & Joint & CLIP-I & DINO \\
    \hline
     \textbf{Supervisor} & & & & \\
    \hline
    SDXL & $\checkmark$ &   &0.7616&0.7040 \\
    SD3  & $\checkmark$ &   &0.7501&0.6965 \\
    SD3.5  & $\checkmark$ &  &0.7567&0.7028 \\
    \hline
     \textbf{Student} & & & & \\
    \hline
    LlamaGen (SDXL) & & $\checkmark$ &\underline{0.7713}&0.7176 \\
    LlamaGen (SD3) & & $\checkmark$ &\textbf{0.7773}&\textbf{0.7335} \\
    LlamaGen (SD3.5) & & $\checkmark$ &0.7597&\underline{0.7235} \\
    \hline
  \end{tabular}
  }
  \caption{Quantitative evaluation on multiple subjects-driven image generation. The diffusion supervisors are trained separately for each subject, while AR students show their capability of learning all subjects jointly.
  }
  \label{tab:multi}
\end{table}

\begin{figure*}[t]
  \includegraphics[width=1\linewidth]{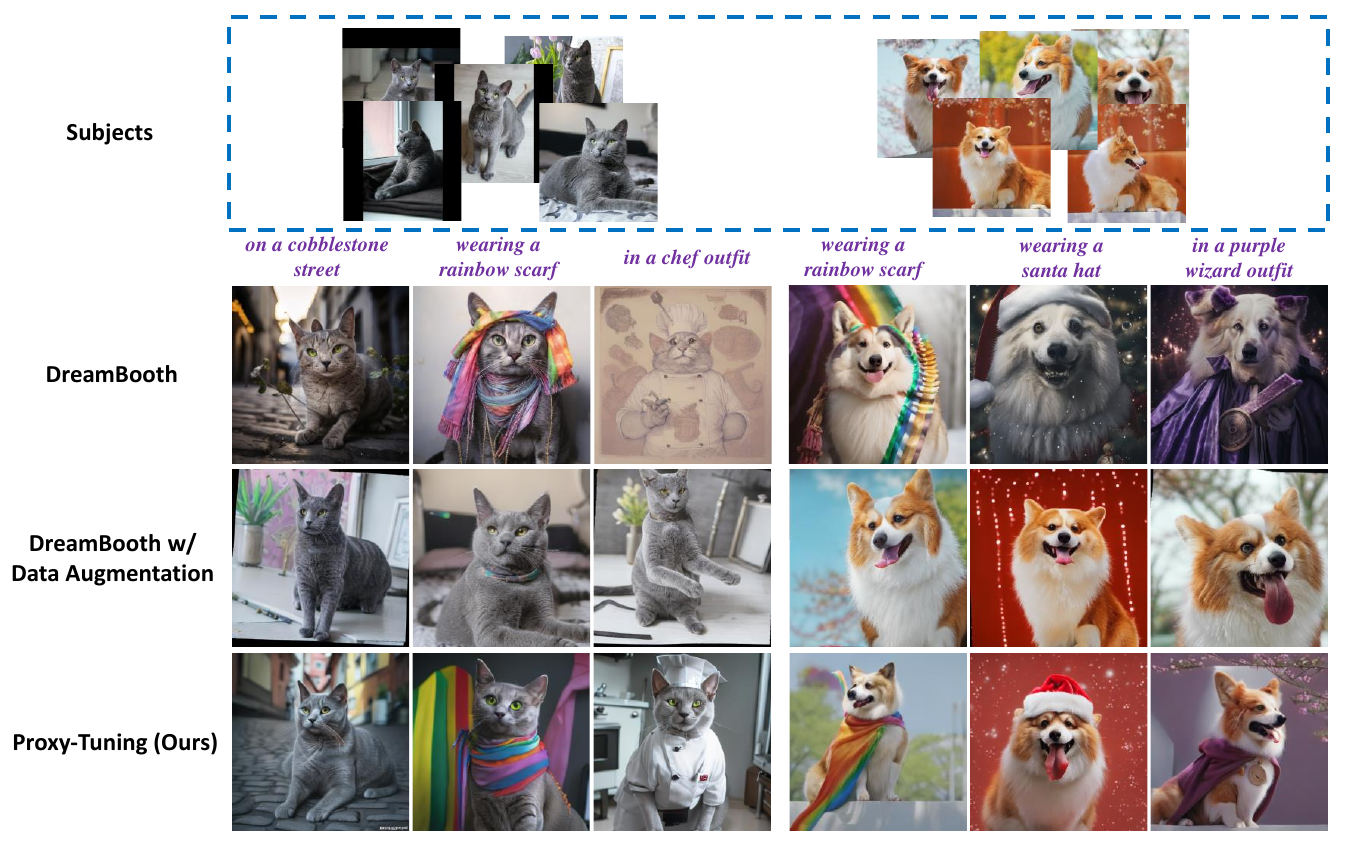}
  \caption {Qualitative comparison between Proxy-Tuning, fine-tuning with data augmentation and fine-tuning with the original dataset. The AR model is LlamaGen and the diffusion supervisor in Proxy-Tuning is SDXL.}
  \label{fig:aug}
\end{figure*}

\subsection{User Study} 
To complement our quantitative metrics and provide human perception-based evaluation, we conduct a user study comparing four approaches: direct subject tuning of diffusion and AR models, Proxy-Tuning of diffusion models, and our proposed Proxy-Tuning of AR models. Concretely, the diffusion model is SDXL and the AR model is Lumina-mGPT. Participants evaluate generated images based on three criteria: image quality, subject fidelity (how well the images preserve the subject's characteristics), and prompt adherence (how well the images align with the given prompts). The study involves 9 participants. The participants are asked to grade the subject images based on three criteria: \textit{image quality}, \textit{subject fidelity}, and \textit{prompt fidelity}. A comprehensive grading is required for the several groups of images of the four methods in the context of three cases. Each case is composed of a subject image and several groups of images obtained from four methods. In each group, there is one piece of text and two images. The results are rated on a scale of 1 to 5, with 5 being the highest. We include direct subject tuning of diffusion and AR models, Proxy-Tuning of diffusion models, and our proposed Proxy-Tuning of AR models. 

The results in Table~\ref{tab:user}, not only align with most of quantitative findings but also reveal an important insight regarding prompt adherence. While CLIP-T scores suggest potential degradation in prompt fidelity with Proxy-Tuning (see Section~\ref{sec:comparison}), human evaluators consistently rate our proxy-tuned AR models higher in prompt adherence. This discrepancy highlights the limitations of CLIP-T as a metric and shows our method's actual performances in maintaining prompt fidelity are better than what automated metrics might suggest.

\begin{table}
  \centering
    \resizebox{.47\textwidth}{!}{
  \begin{tabular}{cccc}
    \hline
     & Image Quality & Subject Fidelity & Prompt Fidelity \\
    \hline
    \textbf{Fine-tuning} \\
    \hline
    Diffusion     &\underline{4.185}  &\underline{3.963}      &\underline{4.370}      \\
    AR    &1.741&1.704&2.926 \\
    \hline 
    \textbf{Proxy-Tuning} \\
    \hline 
    Diffusion     &3.667  & 3.926   & 3.667     \\
    \textbf{AR}    &\textbf{4.259}  &\textbf{4.593}      & \textbf{4.481}     \\
    \hline
  \end{tabular}
  }
  \caption{User study. The compared methods include fine-tuning diffusion model and AR model with original datasets, and Proxy-Tuning diffusion model and AR model on the supervision of weak diffusion supervisor.}
  \label{tab:user}

\end{table}

\subsection{Ablation Study}
\label{sec:ablation}
We explore (1) the impact of supervisor-generated image quantity, (2) the effectiveness compared to traditional data augmentation methods, and (3) the sensitivity to LoRA hyperparameters. Throughout these experiments, we use SDXL as the diffusion supervisor and LlamaGen as the AR student unless otherwise specified. 

\begin{figure}[htbp]
\centering
  \includegraphics[width=1\linewidth]{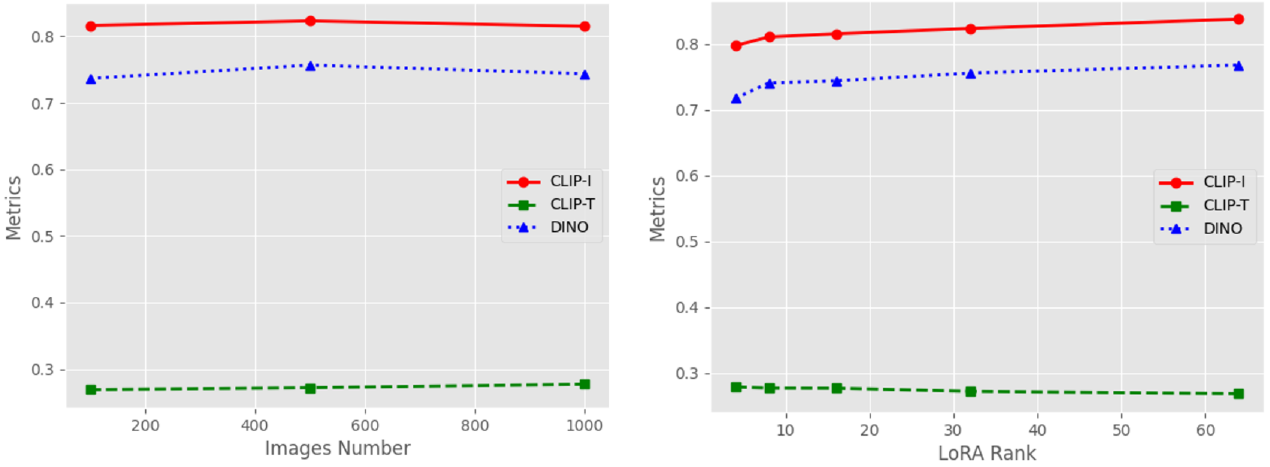}
  \caption{Visualization on ablation of the number of generated proxy images in Proxy-Tuning.}
  \label{fig:aba}
\end{figure}

\noindent\textbf{Impact of Proxy Images.} 
To understand how the quantity of supervisor-generated images affects performance, we experiment with proxy datasets of varying sizes. As shown in Figure~\ref{fig:aba}, Proxy-Tuning demonstrates robust performance across different dataset sizes, suggesting that our method can maintain effectiveness even with a relatively small number of supervisor-generated images.


\noindent\textbf{Data Augmentation.} A natural question is whether data augmentation techniques could achieve similar benefits as our method. We compare Proxy-Tuning with geometric data augmentations applied to the original training images. Specifically, the geometric data augmentations include random horizontal and vertical flipping with a strength of 0.05, random rotation with a strength from -5°to 5°, and random cropping with a strength from 0.9 to 1.0. While these augmentations preserve the target's texture and color characteristics, Figure~\ref{fig:aug} shows that they lead to overfitting and limited semantic editability. This comparison reveals a key insight: AR models' inherent capabilities in semantic understanding and generalization enable them to learn from the rich, contextual variations provided by diffusion supervisors, rather than just surface-level transformations. The superior performance of Proxy-Tuning over data augmentation suggests that AR models can effectively leverage their pre-trained knowledge of semantic relationships to extract and generalize subject-specific features, contributing to the weak-to-strong generalization phenomenon we observed. 

\noindent\textbf{Rank of LoRA.} We investigate the sensitivity of Proxy-Tuning to different LoRA ranks. As shown in Figure~\ref{fig:aba}, our method maintains stable performance across various rank settings, consistently outperforming the subject tuning (LlamaGen) shown in Table~\ref{tab:comparison} and~\ref{tab:comparison_1}. This robustness to LoRA rank selection further demonstrates the reliability of our approach in practical applications.


\section{Conclusion and Limitations}
\label{sec:conclusion}

This paper addresses subject-driven image generation using autoregressive (AR) models. While direct fine-tuning AR models fails to capture subject-specific characteristics, our Proxy-Tuning leverages diffusion models as supervisors to enable effective personalization. Beyond demonstrating robustness across various training configurations, our method reveals a surprising weak-to-strong generalization phenomenon: AR models consistently surpass their diffusion supervisors in both subject fidelity and prompt adherence. Furthermore, AR models exhibit superior parameter efficiency by learning multiple subjects simultaneously, where diffusion models struggle. These advantages establish AR models, when properly guided through Proxy-Tuning, as a scalable and powerful solution for widespread multimodal applications. However, several aspects remain to be explored. First, a theoretical understanding of the weak-to-strong phenomenon deserves deeper investigation, which could provide insights into the fundamental advantages of AR architectures in image generation. Second, our current approach requires empirical selection of proxy dataset sizes; developing automated strategies for determining optimal proxy data quantity could further improve the method's efficiency.

\noindent\textbf{Acknowledgements.} This research is supported by the Anhui Provincial Natural Science Foundation (Grant No.2408085QF214), the Fundamental Research Funds for the Central Universities (Grant No.WK2100000045), the Opening Project of the State Key Laboratory of General Artificial Intelligence (Grant No.SKLAGI2024OP10, Grant No.SKLAGI2024OP11) and the grants from the National Natural Science Foundation of China (No. 62201483).
{
    \small
    \bibliographystyle{ieeenat_fullname}
    \bibliography{main}
}


\end{document}